\newif\ifdraft\draftfalse

\newif\ifinlineref\inlinereffalse
\inlinereftrue

\newif\ifrevmarkerA\revmarkerAfalse

\documentclass[oneside,a4paper]{article}

\usepackage[sort]{natbib}
\usepackage{fullpage}

\usepackage[T1]{fontenc}
\usepackage[utf8]{inputenc}
\usepackage{amsmath}
\usepackage{amsthm}
\usepackage{amsfonts}
\usepackage{setspace}
\usepackage{graphicx}
\usepackage{url}
\usepackage{amssymb}
\usepackage{multirow}
\usepackage{paralist}
\usepackage{listings}
\usepackage{tabularx}
\usepackage{rotating}
\usepackage{booktabs}
\usepackage{tikz}
\usetikzlibrary{shapes,fit,arrows,calc,positioning,shadows}
\usepackage{subcaption}

\usepackage[vlined,algoruled,linesnumbered]{algorithm2e}
\usepackage{fancyvrb}
\usepackage{relsize}
\fvset{fontsize=\relsize{-1},formatcom={\normalfont\ttfamily},numbers=left,numbersep=3pt,xleftmargin=-5pt,numberblanklines=false,frame=single,commentchar=^}
\VerbatimFootnotes

\makeatletter
\renewcommand\verbatim@font{\normalfont\ttfamily}
\makeatother
\renewcommand{\texttt}[1]{\text{\normalfont\ttfamily#1}}

\ifdraft
  \usepackage{todonotes}
\else
  \usepackage[disable]{todonotes}
\fi

\ifrevmarkerA
  \newcommand{\revA}[1]{{\color{blue}#1}}
  \newcommand{\revAc}[0]{\color{blue}}
\else
  \newcommand{\revA}[1]{{#1}}
  \newcommand{\revAc}[0]{}
\fi

\newtheorem{example}{Example}

\newcommand{\leanparagraph}[1]{\smallskip\noindent\textbf{#1.}}

\newcommand{\quo}[1]{``#1''}
\newcommand{\bi}{\begin{itemize}}
\newcommand{\ei}{\end{itemize}}
\newcommand{\bci}{\begin{compactitem}}
\newcommand{\eci}{\end{compactitem}}
\newcommand{\bipe}[1]{\begin{inparaenum}[#1]}
\newcommand{\eipe}{\end{inparaenum}}
\newcommand{\ba}{\begin{array}}
\newcommand{\ea}{\end{array}}
\newcommand{\beq}{\begin{equation}}
\newcommand{\eeq}[1]{\label{#1}\end{equation}}
\newcommand{\mi}[1]{\ensuremath{\mathit{#1}}}
\newcommand{\mc}[1]{\ensuremath{\mathcal{#1}}}
\newcommand{\mb}[1]{\ensuremath{\mathbf{#1}}}

\newcommand{\ins}{\,{\in}\,}
\newcommand{\notins}{\,{\notin}\,}

\newcommand{\ges}{\,{\ge}\,}
\newcommand{\les}{\,{\le}\,}
\newcommand{\lts}{\,{<}\,}
\newcommand{\eqs}{\,{=}\,}
\newcommand{\pluss}{\,{+}\,}

\newcommand{\mids}{\,{\mid}\,}

\newcommand{\modelss}{\,{\models}\,}
\newcommand{\notmodelss}{\,{\not\models}\,}
\newcommand{\cups}{\,{\cup}\,}

\newcommand{\subseteqs}{\,{\subseteq}\,}

\newcommand{\naf}{\texttt{not}}

\newcommand{\cC}{\mc{C}}

\newcommand{\cT}{\mc{T}}
\newcommand{\cX}{\mc{V}}

\newcommand{\HBP}{\mi{H\!B}_P}
\newcommand{\bbN}{\ensuremath{\mathbb{N}}}

\newcommand{\grnd}{\ensuremath{\mi{grnd}}}
\newcommand{\fP}{fP}
\newcommand{\AS}{\mb{AS}}

\newcommand{\clasp}{\textsc{Clasp}}

\newcommand{\clingo}{\textsc{Clingo}}
\newcommand{\gringo}{\textsc{Gringo}}

\newcommand{\ilasp}[0]{\textsc{Ilasp}}
\newcommand{\xhail}[0]{\textsc{Xhail}}
\newcommand{\iled}[0]{\textsc{Iled}}
\newcommand{\mil}[0]{\textsc{Mil}}
\newcommand{\ilaspm}[1]{\ilasp$_{#1}$}
\newcommand{\inspire}[0]{\textsc{Inspire}}
\newcommand{\inspiremn}[0]{\textsc{Inspire}$^{N\text{+}}$}
\newcommand{\inspiremi}[0]{\textsc{Inspire}$^{I\text{+}}$}
\newcommand{\inspireln}[0]{\textsc{Inspire}$^{N\text{-}}$}
\newcommand{\inspireli}[0]{\textsc{Inspire}$^{I\text{-}}$}
\newcommand{\inspirelili}[0]{\textsc{Inspire}$^{I\text{-}\,\text{-}}$}
\newcommand{\inspireliln}[0]{\textsc{Inspire}$^{I\text{-}\,N\text{-}}$}

\newcommand{\Phypgen}{\mi{P_\mi{hypgen}}}
\newcommand{\climitmin}[0]{\ensuremath{\mi{climit}_\mi{min}}}
\newcommand{\climitmax}[0]{\ensuremath{\mi{climit}_\mi{max}}}

\usepackage{myfigures}

\title{Best-Effort Inductive Logic Programming via \\
Fine-grained Cost-based Hypothesis Generation\\[0.5ex]
{\relsize{-2} The Inspire System at the Inductive Logic Programming Competition}}

\author{Peter Sch\"{u}ller$^1$ and Mishal Benz$^2$ \\[1ex]
$^1$
Technische Universit\"at Wien, Institut f\"ur Logic and Computation, Vienna, Austria \\
Marmara University, Faculty of Engineering, Istanbul, Turkey \\
{\tt ps@kr.tuwien.ac.at} / {\tt schueller.p@gmail.com} \\
$^2$
Karlsruhe Institute of Technology, Germany \\
Sabanci University, Faculty of Engineering and Natural Sciences, Istanbul, Turkey \\
{\tt mishal.benz@kit.edu}}

\date{\small%
~\\[1ex]
Technical Report: manuscript submitted to
\emph{Machine Learning} (Springer).}

\begin{document}

\maketitle

\begin{abstract}
We describe the Inspire system which participated in the first
competition on Inductive Logic Programming (ILP).  Inspire is
based on Answer Set Programming (ASP). The distinguishing feature
of Inspire is an ASP encoding for hypothesis space generation: given a set
of facts representing the mode bias, and a set of cost
configuration parameters, each answer set of this encoding
represents a single rule that is considered for finding a
hypothesis that entails the given examples.  Compared with
state-of-the-art methods that use the length of the rule body as
a metric for rule complexity, our approach permits a much more
fine-grained specification of the shape of hypothesis candidate
rules. The Inspire system iteratively
increases the rule cost limit and thereby increases the search space
until it finds a suitable hypothesis. The system searches
for a hypothesis that entails a single example at a time,
utilizing an ASP encoding derived from the encoding used in XHAIL.
We perform experiments
with the development and test set of the ILP competition.  For
comparison we also adapted the ILASP system to process
competition instances.  Experimental results show
that the cost parameters for the
hypothesis search space are an important factor for finding
hypotheses to competition instances within tight resource bounds.

\end{abstract}

\section{Introduction}
\label{secIntro}

Inductive Logic Programming (ILP) \citep{Muggleton2015predinv}
combines several desirable properties
of Machine Learning and Logic Programming:
logical rules are used to formulate
\emph{background knowledge},
and \emph{examples},
which are reasoning inputs
paired with desired or undesired reasoning outcomes,
are used to learn a \emph{hypothesis}.
A hypothesis is an interpretable set of logical rules
which entails the examples with respect to
the background knowledge.
Examples can be noisy, sometimes not all examples
can be satisfied, and usually there are several
possible hypotheses.

The inaugural competition on Inductive Logic Programming
\citep{ILPCompetition} featured a family of ILP tasks
about agents that are moving in a grid world.
Each instance required to find a hypothesis
that represents the rules for valid moves of the agent.
Some instances required predicate invention,
i.e., finding auxiliary predicates
that represent intermediate concepts.
For example the `Unlocked' instance
required the ILP system to find rules for representing
that `the agent may move to an adjacent cell
so long as it is unlocked at that time.
A cell is unlocked if it was not locked at the start,
or if the agent has already visited the key for that cell.'
The competition was open to entries for systems
based on Prolog~\citep{Prolog} and
for systems based on Answer Set Programming (ASP)~%
\citep{Lifschitz2008,Brewka2011cacm,Gebser2012aspbook}
and featured a non-probabilistic and a probabilistic track.

In this paper we describe the \inspire\ system
which is based on ASP and was the winner
of the non-probabilistic competition track,
but it was the only entry to that track.
The competition was challenging for three main reasons.
\begin{enumerate}[(C1)]
\item
  \revA{%
  In each instance the examples
  which were traces of agent movements
  used overlapping time ranges
  and the background knowledge contained
  time comparisons over all earlier time points.
  Therefore, the wide-spread approach of
  shifting the time parameter to represent
  each examples in a distinct part of the Herbrand Base
  was not possible.}%
  \footnote{\revA{To illustrate this, consider the rule
  \quo{\texttt{visited(C, T) :- agent\_at(C, T2), time(T), T >= T2.}}
  which is part of the background knowledge of instance 17 of the competition.
  If we represent multiple sequences of \texttt{agent\_at($\cdot,\cdot$)}
  by allocating time points $0\ldots199$
  for the first and time points $200\ldots{}399$
  for the second agent,
  then the truth values of atoms of form
  \texttt{visited($\cdot,\cdot$)} for the second agent
  will be influenced by the truth values of atoms
  of form \texttt{agent\_at($\cdot,\cdot$)}
  of the first agent.}}%
\item
  Computational resources were limited to 30~sec and 2~GB,
  which is not much for the intractable ILP task.
\item
  Negative example information was given implicitly,
  i.e., agent movements that were not explicitly given
  as valid had to be considered invalid for learning.
\end{enumerate}
Also, at the time of the competition,
there were no published systems that supported
the competition format without the need for
significant adaptations.
\medskip

The \inspire\ system aims to provide a best-effort solution
under these conditions.
The central novel aspect of our approach is
that we generate the hypothesis search space
using an ASP encoding that permits a
fine-grained cost configuration.
We use ASP for
all nontrivial computational tasks
as shown in the block diagram of our system
in Figure~\ref{figFlowchart}.
\flowchart{tbp}

The idea to iteratively extend the hypothesis search space
(in short \emph{hypothesis space})
is present in several existing systems.
\revA{%
Our approach of fine-grained cost-based hypothesis generation
enables a detailed configuration of rule cost parameters,
for example to configure cost
for the number of negative body atoms,
for variables that are bound only once in the rule body,
for invented predicates that are used in the rule body,
for the variables that are bound only in the rule head,
and for several further rule properties.
This provides more control and a more realistic search space
than the common approach of
limiting the rule complexity
which is measured by counting the number of body literals of a rule.
Our approach can be integrated with all ILP systems
that first generate a hypothesis space from the mode bias
and afterwards search for a hypothesis within that hypothesis space.

According to official competition result,
our system predicted 46\% of test cases correctly.
We performed empirical experiments to investigate reasons
for this low accuracy.
Increasing the time budget to 10~min
increases accuracy on test instances by 18\%.
We identify learning from a single example at a time
as a major reason for wrong predictions.
This limitation is due to our hypothesis optimization method
which is derived from the one of \xhail\ and cannot represent multiple examples that share ground atoms,
i.e., it cannot deal with challenge (C1).
In a general setting,
our fine-grained hypothesis search space
is compatible with learning from multiple examples.
\medskip

We make the following contributions.
\begin{itemize}
\item
  We describe an ASP encoding for generating the hypothesis
  search space.
  The encoding permits to attach costs to
  various aspects of rule candidates.
  This way the search space exploration can be controlled
  in a fine-grained way by incrementing a cost limit,
  and preferences for the shape of rule candidates
  can be configured easily.
\item
  We give an algorithm that uses this encoding
  to generate the hypothesis space and
  learns hypotheses from a single example at a time
  using a simplification of the \xhail\ \citep{Ray2009}
  ASP encoding.
  Each hypothesis is validated on all examples
  and if the validation score increased since the last
  validation, a prediction attempt is made,
  followed by hypothesis learning on the next example.
  The algorithm is specific to the competition
  and mainly designed to deal with challenge (C2),
  i.e., obtaining a reasonable score within tight resource bounds.
  The algorithm is based on the observation that
  a single competition example
  often contained enough structure
  to learn the full hypothesis.
\item
  We experimentally compare different cost configurations
  of the \inspire\ system,
  and we compare our system with
  the \ilasp\ system \citep{Law2014ilasp}.
  (For that we created a wrapper to adapt \ilasp\ to
  the competition format and to perform predictions.)
  Our evaluations show that \inspire\ consistently
  outperforms \ilasp,
  that there are significant score differences
  among \inspire\ cost configurations,
  and that learning from single examples
  is not sufficient for all competition instances.
\end{itemize}
}

In Section~\ref{secPrelims}
we provide preliminaries of ASP and ILP
and we describe the ILP format.
In Section~\ref{secHypgen}
we introduce our hypothesis space generation approach%
\revA{, comprising several ASP modules.}
Section~\ref{secBestEffort}
describes the \inspire\ system's algorithm.
The empirical evaluation is reported in Section~\ref{secEvaluation}.
We discuss related work in Section~\ref{secRelated}
and conclude in Section~\ref{secConclusion}.

\section{Preliminaries}
\label{secPrelims}

\subsection{Answer Set Programming}
ASP is a logic programming paradigm
which is suitable for knowledge representation
and finding solutions for computationally (NP-)hard problems~%
\citep{Gelfond1988,Lifschitz2008,Brewka2011cacm}.
We next give preliminaries of ASP programs
with uninterpreted function symbols,
aggregates and choices.
For a more elaborate description we refer to the
ASP-Core-2 standard \citep{Calimeri2012}
and to books about ASP \citep{Baral2004,Gelfond2014aspbook,Gebser2012aspbook}.

\leanparagraph{Syntax}
Let $\cC$ and $\cX$ be mutually disjoint sets
of \emph{constants} and \emph{variables},
which we denote with first letter in lower case and upper case,
respectively.
Constants are used for constant terms, predicate names,
and names for uninterpreted functions.
The set of \emph{terms} $\cT$ is recursively defined:
$\cT$ is the smallest set containing $\bbN \cups \cC \cups \cX$
as well as tuples of form \texttt{($t_1,\ldots,t_n$)}
and uninterpreted function terms of form \texttt{f($t_1,\ldots,t_n$)} where
$\texttt{f} \ins \cC$ and $t_1,\ldots,t_n \ins \cT$.
An \emph{ordinary atom} is of the form \texttt{p($t_1,\dots,t_n$)},
where $\texttt{p} \ins \cC$, $t_1,\dots, t_n \ins \cT$, and
$n\geq 0$ is the \emph{arity} of the atom.
An \emph{aggregate atom} is of the form
\texttt{X\,=\,$\mi{\#agg}$\,\{\,$t$\,:\,$b_1,\,\ldots,\,b_k$\,\}}
with variable $\texttt{X} \ins \cX$, aggregation function
$\mi{\#agg} \ins \{ \texttt{\#sum}, \texttt{\#count} \}$,
with $1 \lts k$,
$t \ins \cT$ and $b_1,\ldots,b_k$ a sequence of atoms.
A term or atom is \emph{ground} if it contains no sub-terms
that are variables.
A \emph{rule $r$} is of the form
$\alpha$ \texttt{:-} $\beta_1,\dots,\beta_n$, \naf\,$\beta_{n+1},\dots$, \naf\,$\beta_{m}$,
where $m\ges 0$, $\alpha$ is an ordinary atom,
$\beta_j$, $0 \les j \les m$ is an atom,
and we let
$B(r) \eqs \{ \beta_1, \dots, \beta_n$,
\naf\,$\beta_{n\pluss 1},\dots$, \naf\,$\beta_m \}$
and
$H(r) \eqs \{ \alpha \}$.
A \emph{program} is a finite set $P$ of rules.
A rule $r$ is a \emph{fact} if $m \eqs 0$.

\leanparagraph{Semantics}
Semantics of an ASP program $P$ is defined
using its Herbrand Base $\HBP$ and its
ground instantiation $\grnd(P)$.
\revA{%
Given an interpretation $I \subseteqs \HBP$
and an atom $a \ins \HBP$,
$I$ models $a$,
formally $I \modelss a$,
iff $a \ins I$
and $I$ models a literal $\naf\,a$,
formally $I \modelss \naf\,a$,
iff $a \notins I$.
}%
An aggregate literal in the body of a rule
accumulates truth values from a set of atoms,
for example $I\ \models\ \texttt{N\,=\,\#count\,\{\,A\,:\,p(A)\,\}}$
iff the extension of predicate \texttt{p} in $I$,
\revA{%
i.e., the set of true atoms of form \texttt{p($\cdot$)},
has size $N$.
}%
\revA{%
An interpretation $I \subseteqs \HBP$
models a rule $r$ if $I \modelss B(r)$ or $I \notmodelss H(r)$,
and $I$ models a set of literals if $I$ models all literals.
}%
The FLP-reduct~\citep{Faber2011} $\fP^I$
reduces a program $P$ using an answer set candidate $I$:
$\fP^I \eqs \{ r \ins \grnd(P) \mids I \modelss B(r) \}$.
Finally
$I$ is an answer set of $P$, denoted $I \ins \AS(P)$,
iff $I$ is a minimal model of $\fP^I$.

\leanparagraph{Syntactic Sugar}
Anonymous variables of form \quo{$\_$}
are replaced by new variable symbols.
Choice constructions can occur instead of rule heads,
they generate a set of candidate solutions if the rule body is satisfied;
e.g., \texttt{1\,\{\,p(a)\,;\,p(b)\,\}\,2} in the rule head
generates all solution candidates where at least 1 and at most 2 atoms of the set $\{\texttt{p(a)$,\,$p(b)}\}$ are true (bounds can be omitted).
If a term is given as $X\texttt{..}Y$, where $X,Y \ins \bbN$,
then the rule containing the term is instantiated with all
values from $\{ v \ins \bbN \mids X \les v \les Y \}$.
\revA{%
A \emph{constraint} is a rule $r$ without a head atom,
and a constraint eliminates answer sets $I$ where $I \models B(r)$.
A constraint can be rewritten into a rule \texttt{f\,:-\,\naf\,f,\,$B(r)$},
where $f$ is an atom that does not occur elsewhere in the program.}

ASP supports optimization
by means of weak constraints which incur a cost
instead of eliminating an answer set.
We denote by $\AS^{\mi{opt},1}(P)$
the first optimal answer set of program $P$.
Note, that the hypothesis space generation encodings,
which are the main contribution of this paper,
do not require weak constraints
because they explicitly represent costs.

\subsection{Inductive Logic Programming}

ILP \citep{muggleton1994inductive,Muggleton2012ilp20}
is a combination of Machine Learning with logical knowledge representation.
Key advantages of ILP are the generation of \revA{compact models
that can be interpreted by humans,
and the possibility to learn from a small amount of examples}.
A classical ILP system takes as input a set of examples $E$,
a set $B$ of background knowledge rules, and
a set of mode declarations $M$, also called the mode bias.
An ILP system is expected to produce a set of rules $H$ called the hypothesis
which entails $E$ with respect to $B$
in the underlying logic programming formalism.
The search for $H$ with respect to $E$ and $B$ is restricted by $M$,
which defines a language that limits the shape of rules
that can occur in the hypothesis.

Traditional ILP \citep{Muggleton2012ilp20} searches for sets of Prolog rules
\revA{that entail a given set of positive examples
and do not entail a given set of negative examples} using SLD resolution
and a given set of background theory rules.
\revA{Brave Induction \citep{Sakama2009braveinduction}
requires that each example is entailed in at least one answer set,
while cautious induction requires all examples to be entailed in all answer sets.}
ILP for ASP was introduced by \cite{Otero2001aspilp}
and searches for a set of ASP rules
that \revA{entails each given example (consisting of a positive and negative part)
in at least one answer set}.
ASP hypotheses represent knowledge in a more declarative way than Prolog,
i.e., without relying on the SLD(NF) algorithm.

\begin{example}
Consider the following example ILP instance $(M,E,B)$ \citep{Ray2009}.
\begin{align*}
&M = \left \{
  \texttt{%
  \begin{tabular}{@{}l@{}}
  \#modeh flies(+bird).\\
  \#modeb penguin(+bird).\\
  \#modeb \naf\ penguin(+bird).
  \end{tabular}%
  }
\right \}
\\
&E = \left \{
  \texttt{%
  \begin{tabular}{@{}l@{}}
  \#example flies(a). \\
  \#example flies(b). \\
  \#example flies(c). \\
  \#example \naf\ flies(d).
  \end{tabular}%
  }
\right \}
\\
&B = \left \{
  \texttt{%
  \begin{tabular}{@{}l@{}}
  bird(X) :- penguin(X). \\
  bird(a). \\
  bird(b). \\
  bird(c). \\
  penguin(d).
  \end{tabular}%
  }
\right \}
\\
\intertext{%
Based on the above, an ILP system would ideally find the following hypothesis.
}%
  &H = \left \{
    \texttt{%
    \begin{tabular}{@{\,}l@{\,}}
    flies(X) :- bird(X), \naf\ penguin(X).
    \end{tabular}%
    }
  \right \}
\end{align*}
\end{example}
\revA{Note that, in this example,
the program $B \cups H$ has a single answer set
that entails $E$.

There are also ASP-based ILP systems
(for example \ilasp\ \citep{Law2014ilasp}) where
a hypothesis
\revA{must entail each positive examples in some answer set,
and no negative example in any answer set.}
With that, ILP can be used to learn, e.g., the rules of Sudoku:}
given
  a background theory that generates all answer sets of~9-by-9
  grids containing digits~1 through~9,
  positive examples of partial Sudoku solutions, and
  negative examples of partial invalid Sudoku solutions,
ILP methods can learn
which rules define valid Sudoku solutions.
\revA{More recently, ASP-based ILP has been extended to inductive learning of
preference specifications in the form of weak ASP constraints \citep{Law2015ilaspw}.}

\subsection{First Inductive Logic Programming Competition}
\label{secILPCompetition}

The first international ILP competition,
held together with the 26th International Conference on Inductive Logic Programming,
aimed to \revA{``test the accuracy, scalability, and versatility
[of participating ILP systems]'' \citep{ILPCompetition}}.
The competition comprised a probabilistic and a non-probabilistic track.
We consider only the non-probabilistic track here.

The initial datasets consisted of 8 example problems in each track,
intended to help entrants build their systems.
The datasets used for scoring systems in the competition
were completely new and unseen.
All runs were made on an Ubuntu 16.04 virtual machine
with a 2~GHz dual core processor and resource limits of
2~GB RAM and 30~sec time.

Instances were in the domain of agents moving in a grid world
where only some movements were possible.
Each instance consists of a background knowledge, a language bias,
a set of examples, and a set of test traces.
A trace is a set of agent positions at certain time positions.
An example contains a trace and a set of valid moves.
The ILP system had to learn the rules for possible moves
from examples,
and then predict for each test trace
whether the agent made only valid moves.
These predictions were used to produce the final score.

\begin{example}
In \revA{Instance~5, called }\emph{Gaps in the floor},
the agent can always move sideways,
but can only move up or down in special `gap' cells
which have no floor and no ceiling.

In \revA{Instance~11, called }\emph{non-OPL transitive links},
the agent may go to any adjacent cell
or use given links between cells to teleport.
It can also use a chain of links in one go.
In this problem, the agent has to learn the
(transitive) concept `linked'.
\end{example}
Each input instance is structured in sections using the following statements
\revA{see Figure~\ref{figSimpleTask} for an Example}:
\begin{itemize}
\item \verb+#background+ marks the beginning of the background knowledge.
\item \verb+#target_predicate+ indicates the predicate which should be defined by the hypothesis,
 similar to the \emph{modeh} mode declaration in standard language biases.
\item
  \verb+#relevant_predicates+ indicates the predicates from the background knowledge
  that %
  can be used to define the hypothesis,
  similar to
  \emph{modeb} mode declarations in standard language biases.
\item \verb+#Example(X)+ shows the start of an example with identifier \texttt{X}. Subsection \verb+#trace+ contains the path taken by the agent,
and subsection \verb+#valid_moves+ gives the complete set of valid moves the agent could take for each time step.
\item \verb+#Test(X)+ contains a test trace with identifier \texttt{X}.
Subsection \verb+#trace+ contains the path taken by the agent.
\end{itemize}
Predicates in the language bias contained only
variable types as arguments, never constants.

\revA{An ILP system in the competition is supposed to
learn a hypothesis based on the given examples (traces and valid moves) and the background knowledge,
and then predict the valid moves of the given test traces
using the learned hypothesis and the background knowledge.}
The system output consists of answer attempts,
which start with \verb+#attempt+,
followed by a sequence of lines \texttt{VALID($X$)} or \texttt{INVALID($X$)},
predicting validity of agent movements in each test trace $X$.
Multiple answer attempts are accepted,
but only the last one is scored.

\figureSimpleTask
\begin{example}
  \label{exSimpleTask}
  A simplification of Instance 6 \revA{from the competition }%
  is given in Figure~\ref{figSimpleTask}.
\end{example}

\section{Declarative Hypothesis Generation}
\label{secHypgen}
A central part of our ILP approach is,
that we use an ASP encoding to generate
the hypothesis space
which is the set of rules that is considered
to be part of a hypothesis.
Concretely, we represent the mode bias
of the instance as facts,
add them to an answer set encoding
which we describe in the following,
moreover we add a fact that configures the cost limit
for rules in the hypothesis space.
Each answer set represents a single rule of the hypothesis space.
\revA{While this hypothesis space does not permit to find a hypothesis,
we increment the cost limit to enlarge the hypothesis space
until we find a solution.}
\medskip

We use the following representation
for predicate (schemas) $P(t_1,\ldots,t_N)$
with predicate name $P$, arity $N$,
and argument types $t_1, \ldots, t_N$:
\begin{itemize}
\item \texttt{pred($I,P,N$)}
  represents predicate $P$ with arity $N$,
  where $I$ is a unique identifier for this predicate and arity; and%
\item \texttt{arg($I,j,t_j$)}
  represents the type $t_j$ of argument position $j$ of predicate $I$.
\end{itemize}

\begin{example}
  The predicate \verb+valid_move(cell,time)+,
  which was the target predicate in many instances of the competition,
  is represented by the following atoms,
  where \verb+p1+ is the unique predicate identifier.
  \begin{LVerbatim}[gobble=2]
  pred(p1,valid_move,2).
  arg(p1,1,cell).
  arg(p1,2,time).
  \end{LVerbatim}
\end{example}

\subsection{Input Representation}
\label{secHypgenInput}
Hypothesis space generation is based on a
target predicate and relevant predicates of the instance
at hand.
We represent this input in atoms of the following form,
using above schema:
\begin{itemize}
\item \texttt{tpred($I,P,N$)}
  for the target predicate;
\item \texttt{targ($I,J,T$)}
  for arguments of the target predicate;
\item \texttt{rpred($I,P,N$)}
  for relevant predicates;
\item \texttt{rarg($I,J,T$)}
  for arguments of relevant predicates; and
\item \texttt{type\_id($T,\mi{ID}$)}
  for all types $T$ used in the target predicate
  and in relevant predicates, where \mi{ID}
  is the \emph{type identifier},
  a unique integer associated with $T$.
  The set of all type identifiers
  must form a zero-based continuous sequence.
\end{itemize}

\begin{example}
  \label{exSimpleTaskBias}
  \revAc
  The mode bias that is given in Figure~\ref{figSimpleTask}
  is represented as follows.
  The target predicate \texttt{valid\_move(cell,time)}
  is represented by the following facts,
  where \texttt{t1} is an identifier for the predicate,
  and \texttt{1} and \texttt{2} in \texttt{targ} are argument positions of the predicate.
  \begin{LVerbatim}[gobble=2]
  tpred(t1,valid_move,2). targ(t1,1,cell). targ(t1,2,time).
  \end{LVerbatim}
  The relevant predicates \texttt{gap(cell)}, \texttt{agent\_at(cell,time)},
  \texttt{h\_adjacent(cell,cell)}, and \texttt{v\_adjacent(}\allowbreak\texttt{cell,cell)},
  are represented by the following facts,
  where \texttt{r1}, \ldots, \texttt{r4} are the respective predicate identifiers.
  \begin{LVerbatim}[gobble=2,firstnumber=last]
  rpred(r1,gap,1).        rarg(r1,1,cell).
  rpred(r2,agent_at,2).   rarg(r2,1,cell). rarg(r2,2,time).
  rpred(r3,h_adjacent,2). rarg(r3,1,cell). rarg(r3,2,cell).
  rpred(r4,v_adjacent,2). rarg(r4,1,cell). rarg(r4,2,cell).
  \end{LVerbatim}
  Finally, the following facts define type identifiers \texttt{0} and \texttt{1}
  for \texttt{cell} and \texttt{time}, respectively.
  \begin{LVerbatim}[gobble=2,firstnumber=last]
  type_id(cell,0).
  type_id(time,1).
  \end{LVerbatim}
\end{example}

\subsection{Output Representation}
\label{secHypgenOutput}
We represent a single rule per answer set
during hypothesis space generation.

A rule is represented in atoms of the following form:
\begin{itemize}
\item \texttt{use\_var\_type($V,T$)}
  represents that the rule uses variable $V$ with type $T$.
  $V$ is a term of form \texttt{v(\mi{Idx})}
  denoting variable with index $\mi{Idx}$
  and $T$ is a type as provided in
  input atoms as first argument of predicate \texttt{type\_id}
  for relevant predicates.
\item \texttt{use\_head\_pred($\mi{Id,Pred,A}$)}
  represents that the rule head
  is an atom with predicate identifier $\mi{Id}$,
  predicate $\mi{Pred}$, and arity $A$.
\item \texttt{use\_body\_pred($\mi{Id,Pred,Pol,A}$)}
  represents that the rule body contains a literal
  with literal identifier $\mi{Id}$,
  predicate $\mi{Pred}$, of polarity $\mi{Pol}$, and arity $A$.
  Importantly,
  \revA{if a predicate is used in multiple body literals,
  $\mi{Id}$ is different for each literal;
  $\mi{Id}$ is also used in \texttt{bind\_bvar}
  (see below) for binding variables
  to argument positions of particular literals.}
\item \texttt{bind\_hvar($\mi{J,V}$)}
  represents that the argument position $J$
  in the rule head contains variable $V$,
  where $V$ is a term of form $v(\mi{Idx})$.
\item \texttt{bind\_bvar($\mi{Id,Pol,J,V}$)}
  represents that the rule body literal with identifier $\mi{Id}$
  and polarity $\mi{Pol}$ contains in its argument position $J$
  the variable $V$.
  ($\mi{Id}$ refers to a unique body literal
  as represented in the argument $\mi{Id}$
  of \texttt{use\_body\_pred}, see above.)
\end{itemize}

\begin{example}\label{exHypCand}
  The hypothesis candidate
  \begin{LVerbatim}[gobble=2]
  valid_move(V5,V10) :- cell(V5), time(V10), not agent_at(V5,V10).
  \end{LVerbatim}
  is represented in an answer set by the following atoms.
  \begin{LVerbatim}[gobble=2]
  use_var_type(v(5),cell)
  use_var_type(v(10),time)
  use_head_pred(t1,valid_move,2)
  bind_hvar(1,v(5))
  bind_hvar(2,v(10))
  use_body_pred(id_idx(r2,1),agent_at,neg,2)
  bind_bvar(id_idx(r2,1),neg,1,v(5))
  bind_bvar(id_idx(r2,1),neg,2,v(10))
  \end{LVerbatim}
  \revA{As in Example~\ref{exSimpleTaskBias},
  \texttt{t1} represents target predicate \texttt{valid\_move(cell,time)}
  and \texttt{r2} represents the relevant predicate \texttt{agent\_at(cell,time)}.
  Line~1 represents that variable \texttt{V5} has type \texttt{cell} and
  line~2 represents that variable \texttt{V10} has type \texttt{time}.
  Line~3 represents that the rule head contains target predicate \texttt{valid\_move},
  and lines~4 and~5 represent that the first and second argument positions of
  the head atom
  are bound to the variables \texttt{V5} and \texttt{V10}, respectively.
  Lines~6--8 represent the body literal \texttt{\naf\ agent\_at($\cdot$,$\cdot$)}:
  line~6 represents that the body contains
  a negated literal with predicate \texttt{agent\_at},
  and this literal has the unique identifier \texttt{id\_idx(r2,1)};
  lines~7--8 represent that the first and second argument positions
  of this literal
  are bound to variables \texttt{V5} and \texttt{V10}, respectively.}
  The body literals \texttt{cell(V5)} and \texttt{time(V10)}
  are not explicitly represented,
  they are implicit from
  \texttt{use\_var\_type}.%
\end{example}

\subsection{Cost Configuration}
\label{secHypgenConfig}

For fine-grained control over the
shape of rules in the hypothesis space,
we define several cost components on rules.
\revA{Intuitively,
rules with lower overall cost will be considered in the search space earlier
than rules with higher cost.
All rules below a certain cost are used simultaneously
for finding a hypothesis that entails a given example
(see also Section~\ref{secBestEffort} and Algorithm~\ref{algo}).
\medskip

\textbf{Hard Limits.~}
For ensuring decidability,
it is necessary to impose hard limits
on the overall size of the hypothesis space.}
We use the following hard restrictions on rules
in the hypothesis search space.
\revA{Configuration parameters are written in bold
and default values are given in brackets.}
\begin{itemize}
\item
  $\mb{maxvars}$ (4)
  specifies the maximum number of variables per type.
  This limits how many variables
  of a single type can occur simultaneously in one rule.
\item
  $\mb{maxuseppred}$ (2)
  specifies the maximum occurrence of a single predicate
  as a positive body literal.
  This limits how often we can use the same predicate
  in the positive rule body.
  For example, for obtaining a transitive closure
  in the hypothesis space, this value needs to be
  at least 2, and $\mb{maxvars}$ needs to be at least 3.
\item
  $\mb{maxusenpred}$ (2)
  specifies the maximum occurrence of a single predicate as a negative body literal.
\item
  $\mb{maxliterals}$ (4)
  specifies the maximum number of overall literals in a rule.
  This \revA{imposes a hard limit on the size of }hypothesis rule bodies.
\item
  $\mb{maxinventpred}$ (1)
  specifies the maximum number of predicates to be invented.
\item
  $\mb{inv\_minarity}$ (2)
  specifies the minimum arity of invented predicates.
\item
  $\mb{inv\_maxarity}$ (2)
  specifies the maximum arity of invented predicates.
\end{itemize}
\revA{These hard limits make the hypothesis search space finite
by limiting the usage of the mode bias,
therefore their values must be chosen with care.}
Moreover, these limits determine the size of the instantiation of the ASP encoding,
which influences the efficiency of enumerating answer sets.
\revA{In Section~\ref{secSoundComplete} we discuss soundness and completeness of the \inspire\ system with respect to this finite search space.
\medskip

\textbf{Fine-grained cost configuration.~}
For configuring fine-grained hypothesis generation,
we provide the following cost parameters
for various aspects of rules in  the hypothesis search space.
(Defaults are again given in brackets.)}%
\begin{itemize}
\item
	$\mb{free\_vars}$ (2)
	specifies the
  number of variables that do not incur cost.
\item
	$\mb{cost\_vars}$ (1)
	specifies the
  cost for each variable beyond $\mb{free\_vars}$.
\item
	$\mb{cost\_type\_usedmorethantwice}$ (2)
	specifies the
  cost for each usage of a type beyond the second usage.
  For example, this cost is incurred
  if we use three variables of type \texttt{time}
  in one rule.
\item
	$\mb{cost\_posbodyliteral}$ (1)
	specifies the
  cost for each positive body literal.
\item
	$\mb{cost\_negbodyliteral}$ (2)
	specifies the
  cost for each negative body literal.
  The default value is higher than the one for
  $\mb{cost\_posbodyliteral}$,
  because usually programs have more positive
  body literals than negative body literals.
\item
	$\mb{cost\_pred\_multi}$ (2)
	specifies the
  cost for repeated usage of a predicate in the rule body.
  The cost is incurred for each usage after the first usage.
  Usage is counted separately for positive and negative literals.
\item
	$\mb{cost\_varonlyhead}$ (5)
	specifies the
  cost for each variable that is used only in the head.
  Rules with such variables are still safe because each variable
  has a type.
  It is possible, but rare, that such rules are useful,
  so they obtain high cost.
\item
	$\mb{cost\_varonlyoncebody}$ (5)
	specifies the
  cost for each variable that is not used in the head
  and used only once in the body of the rule.
  Such variables are like anonymous variables
  and project away the argument where they \revA{occur}.
  We expect such cases to be rare
  so we incur a high default cost.
\item
	$\mb{cost\_var\_boundmorethantwice}$ (2)
	specifies the
  cost for each variable that \revA{occurs }in more than two literals.
\item
	$\mb{cost\_reflexive}$ (5)
	specifies the
  cost for each binary atom in the rule body with the same variable in both arguments.
\item
	$\mb{cost\_inv}$ (2)
	specifies the
  cost for inventing any \revA{kind }of predicate.
  This cost is used to adjust above which cost predicate invention
  is performed, independent from the cost of inventing each predicate.
\item
	$\mb{cost\_inv\_pred}$ (2)
	specifies the
  cost for each invented predicate.
\item
	$\mb{cost\_inv\_headbody}$ (3)
	specifies the
  cost for using the same invented predicate both in the head and in the body of the same rule.
\item
	$\mb{cost\_inv\_bodymulti}$ (5)
	specifies the
  cost for using multiple invented predicates in the rule body.
\item
	$\mb{cost\_inv\_headbodyorder}$ (5)
	specifies the
  cost for using an invented predicates
  in the head and a different invented predicate
  in the body of a rule,
  in a way that the head predicate is
  lexically greater than the body predicate.
  This incurs higher cost to programs with cycles
  over invented predicates,
  and less cost to those that have no such cycles.
  In particular this allows for an early consideration of hypotheses
  that use invented predicates in a \quo{stratified} way
  such that they have no chance to introduce nondeterminism
  (by means of even loops over invented predicates).
\end{itemize}
The above costs can independently be adjusted
and influence the performance of our approach.
Costs determine the stage of the ILP search
at which a certain rule will be used as a candidate
for the hypothesis.

\revA{If we expect certain rules to be more useful
for finding a hypothesis in a concrete application,
they should be configured to have lower cost
than other rules.}

\figureSimpleHypspace
\begin{example}
  \label{exHypSpace}
  The hypothesis candidate shown in Example~\ref{exHypCand}
  has a cost of 2 using the default cost settings,
  because of one cost component from \mb{cost\_negbodyliteral}.
  \revA{%
  Figure~\ref{figSimpleHypspace} shows all rules of cost 1--5
  for the instance given in Figure~\ref{figSimpleTask}
  using default cost parameters.
  In (i), only the cost parameter $\mb{cost\_posbodyliteral}=1$ is effective.
  Similarly, in (ii), only $\mb{cost\_negbodyliteral}=2$ has an effect.
  In (iii), for a total cost of $3$,
  each candidate obtains cost $\mb{cost\_vars}=1$ for using three variables
  (two variables incur no cost because $\mb{free\_vars}=2$)
  and each candidate obtains cost $2$ for two positive body literals.
  In (iv), for a total cost of $4$,
  the first rule obtains cost $\mb{cost\_var\_boundmorethantwice}=2$
  for \texttt{V5} which is used three times in rule head and rule body,
  moreover cost $2$ for two positive body literals.
  Each of the remaining eight rules in (iv) obtains $\mb{cost\_vars}=1$ for using three variables,
  cost $1$ for a positive body literal, and cost $2$ for a negative body literal.
  In (v), for a total cost of $5$,
  candidates in lines~1--2 obtain cost $\mb{cost\_var\_boundmorethantwice}=2$
  for using \texttt{V5} three times and cost for one positive and one negative body literal.
  Candidates in lines~3--4 obtain cost $\mb{cost\_vars}=1$ for using three variables,
  cost $2$ for two positive body literals, and cost $2$ for one negative body literal.
  Candidates in lines~5--8 obtain cost $\mb{cost\_vars}=1$ for using three variables
  and cost $2$ for two negative body literals.
  Candidates in lines~9--14 obtain cost $\mb{cost\_inv}=2$ for inventing predicates,
  $\mb{cost\_inv\_pred}=2$ for the first invented predicate \texttt{ip\_1\_2},
  and cost $1$ for a positive body literal.
  Note that the shown rules are the actual internal system representation
  where redundant candidates (with renamed variables)
  have been eliminated (see Section~\ref{secCodeRed}).
  The fine-grained nature of this cost configuration becomes apparent when
  considering the \quo{classical} cost notion of rule body length:
  all rules in (i) and (ii) and rule in lines~9--14 of (v) have bodies of length 1,
  rules in lines~2 and~3 of (v) have bodies of length 3,
  and all remaining rules shown in Figure~\ref{figSimpleHypspace} have bodies of length 2.}%
\end{example}

\revA{Different from hard limits,
adjusting cost parameters has no influence on the possibility
to find a hypothesis.
Instead, these parameters intuitively control search heuristics:
they influence in which order
hypotheses are considered and therefore can speed up or slow down
hypothesis search.}

\figureHypgenMain
\subsection{Main Encoding}
\label{secCodeMain}
The main encoding for generating the hypothesis space
is given in Figure~\ref{figCodeMain}.
Hard limits and cost parameters
are added to this encoding as constant definitions.

We define distinct typed variables
in atoms of form \texttt{var\_type(v($\mi{Index}$),$T$)}
in line~1.
Such an atom represents,
that the variable of form \texttt{v(\mi{Index})}
has type $T$,
where $\mi{Index}$ is a running index
over all variables.
This defines \mb{maxvars} variables of each type.
Note that we use \texttt{v(\mi{Idx})} to enable a later extension of our encodings
with constant strings as arguments in the mode bias.
Constant strings were not required for the ILP competition.

Head predicates are represented as \texttt{hpred} and \texttt{harg},
and body predicates are represented as \texttt{bpred} and \texttt{barg}
in lines~2--5.
These are defined from target predicate
and relevant predicate, respectively.
We define further head and body predicates
in the encoding for invented
predicates (see Section~\ref{secCodeInv}).

We guess how many variables
(up to \mb{maxvars}) are in the rule
in the current answer set candidate (line~6).
We guess which concrete variables
(including their type) are in the rule (line~7).
\revA{%
Atoms of form \texttt{use\_var\_type($V,T$)}
represent that variable $V$ has type $T$.
}%

\revA{%
Lines~8--11
define unique placeholders
for all potentially existing literals in the rule body
according to the given hard limits.
}%
Such placeholders
are represented in atoms of form
\verb+body_pred(ID,Pred,Pol,A)+
where $ID$ is a unique term built for that predicate
from its predicate identifier $Id$ and a running index $Idx$,
$Pred$ is the predicate name, $Pol$ the polarity,
and $A$ the arity of the predicate.

A subset of these placeholders is guessed
as a body literal in lines~12--13,
up to a maximum of $\mb{maxliterals}$ literals.
A guess in line~14 determines the head predicate.
\revA{Up to this point, }%
the encoding represents variables including their type,
which variables are going to be used,
and which head and body predicates to use as literals.

\revA{In lines~15--19, we perform a guess
for binding
these variables to particular argument positions
of head and body predicates.}
The limits of these choice rules require
that each position is bound exactly once.
Moreover, the conditions within the choice ensures
that variables are bound to argument positions
of the correct type.
In lines~20--21,
we represent the set of variables that are bound in the
head and the same for the rule body
(this separation is used for cost representations).
Finally, an answer set where a variable
is used but neither bound to the head
nor to the body of the rule,
is eliminated by constraint in line~22.
We do not forbid `unsafe rules'
(where a variable exists only in the head of a rule)
because all variables are typed
and therefore each variable occurs in an implicit domain predicate
in the rule body.

If we evaluate this program module
together with a mode bias given as facts
according to Section~\ref{secHypgenInput}
and together with constant definitions of
hard limits according to Section~\ref{secHypgenConfig},
we obtain answer sets that represent single rules
according to Section~\ref{secHypgenOutput}
and according to the given mode bias and hard limits.

\subsection{Fine-Grained Cost Module}
\label{secCodeCost}%
\figureHypgenCost%
Figure~\ref{figCodeCost} shows the encoding module
for representing the cost of a rule
according to cost configuration parameters
described in Section~\ref{secHypgenConfig}.

Cost atoms of the form
\texttt{cost(\mi{Name,Data,Cost})}
represent various costs that add up to the total rule cost.
Each cost atom bears a name $\mi{Name}$ used to distinguish
different aspects of cost.
\revA{%
The argument $\mi{Data}$ specifies different elements
of the same rule (e.g., variables, literals)
that can contribute cost under that aspect.}
Finally $\mi{Cost}$ is the actual cost incurred for $\mi{Name}$ and $\mi{Data}$.
\begin{example}
\revA{For the aspect of using a variable type more than twice,
the cost aspect is}
$\mi{Name}\ =\ $\texttt{vartype\_more\allowbreak{}than\allowbreak{}twice},
and $\mi{Data}$ contains the variable type for which this cost
is incurred.
\revA{For each variable type, this aspect incurs cost separately,
which leads to multiple atoms with different $\mi{Data}$ values.}
\end{example}

Lines~1--2 define a cost for the number of distinct variables
that are used in the rule, where the first
$\mb{free\_distinct\_variables}$ variables incur no cost.
Lines~3--4 define the cost for variable types that are used more than twice
(using a type once or twice is free).
Lines~5--6 define costs for positive and negative body literals,
and lines~7--8 define costs for using a predicate multiple times in the body.
Note, that line~7 relies on the property that the body literals of lowest index are used,
which is ensured by the redundancy elimination module
(see Section~\ref{secCodeRed}, lines~2--3 in Figure~\ref{figCodeRed}).
Lines~9--10 define a cost for each variable that \revA{occurs }only in the head.
Lines~11--12 define a cost for variables that occur only once in the body of the rule and not in the rule head \revA{(these act as anonymous variables)}.
Lines~13--16 define a cost for variables that occur more than twice in the rule.
Lines~17--21 define a cost for \revA{the reflexive usage of a binary predicate,
i.e., a cost for literals that} contain the same variable in both arguments.
Lines~22--23 sum up the total cost if that total is below $\texttt{climit}$,
otherwise the total cost is fixed to $\texttt{climit}$.%
\footnote{%
The rule \texttt{totalcost(C) :- C = \#sum \{ Cost,U,V : cost(U,V,Cost) \}.}
would sum up total cost in a single rule,
without clamping the value to the maximum interesting value
$\texttt{climit}$.
\revA{%
However, this naive approach has a significantly larger instantiation
than the encoding we use here}.}
Finally, solutions that reach or exceed a total cost
of $\texttt{climit}$ are excluded in line~24.

\figureHypgenRed
\subsection{Redundancy Elimination Module}
\label{secCodeRed}
The encoding in Figure~\ref{figCodeMain}
creates many solutions that produce the same
or a logically equivalent rule.
As an example,
line~1 defines $\mb{maxvars}$ variables of the same type,
and line~7 guesses which of these variables to use.
If variables with index 1 and 2 have the same type,
there can be two answer sets which represent
two rules that are different modulo variable renaming.
For example, one of the rules
\quo{\texttt{foo(V1) :- bar(V1).}}
and
\quo{\texttt{foo(V2) :- bar(V2).}}
is redundant.
\revA{%
Redundant rules make the hypothesis search slower
and do not contribute to the solution,
therefore they should be avoided.
}%

Figure~\ref{figCodeRed} is an ASP module
that eliminates most redundancies
and thereby improves performance
without losing any potential hypotheses.
Line~1 ensures that if we use variables of a certain type,
we use only variables with the lowest index of that type.
This is realized by
\revA{enforcing that variable \texttt{v($Id\text{-}1$)} is used
whenever variable \texttt{v($Id$)} is used
and under the condition that these variables have the same type.}
Similarly, lines~2--3 require that those
body literals that have the lowest indexes are used.
Lines~4--7 canonicalize variables that are used in rule heads:
the constraint in lines~4--5 requires that the variables with lowest index
are bound in the head,
and the constraint in line~6--7 rules out solutions where two variables
of the same type are used in the head
where the variable with the lower index is used
in the second argument of the predicate.
This rules out, e.g., a rule with the head
\verb+foo(X2,X1)+ if \verb+X1+ and \verb+X2+ have the same type.

For further redundancy elimination
we rely on the lexicographic order of terms in ASP.
In lines~8--12
\revA{%
we define atoms which represent a \emph{variable signature} $S$
of form \texttt{lit\_vsig($I$,\allowbreak{}$\mi{Idx},\mi{Pol},A,S$)}
}%
where $I$ is the predicate identifier, $Idx$ the index,
$Pol$ the polarity, $A$ the arity,
and $S$ is a composite term containing all variables
in the literal for which the signature is defined.
Using these signatures,
the constraint in line~13
requires that literals with equal predicates and polarities
are sorted in the same way as their variable signatures.
This would, for example, eliminate a rule with
the body \quo{\texttt{foo(X2,X3), foo(X1,X2)}}
while it would allow to use the logically equivalent
body \quo{\texttt{foo(X1,X2), foo(X2,X3)}}.
\revA{%
Finally, in lines~14--20, we represent pairs of literals
that have the same predicate and the same arguments
(and potentially different polarity),
and we exclude solutions that contain a pair of such equal literals
using the constraint in line~21.}

Note that lines~6--12 of this encoding are suitable only
for predicates with arity one or two.
This was sufficient for the ILP competition
and can be generalized to higher arities.

\figureHypgenInv%
\subsection{Predicate Invention Module}%
\label{secCodeInv}%
Figure~\ref{figCodeInv} shows the ASP module
which extends the search bias
by adding the possibility to invent predicates
in the hypothesis search space.
This module also includes redundancy elimination aspects
that are specific for predicate invention.

Lines~1--2 \revA{define a guess over the arity
of up to} \mb{maxinventpred} invented predicates.
Guessing no arity means that the invented predicate
is not used.
\revA{Line~3 guesses for each invented predicate
and for each argument position one argument type.}
Lines~4--7 connect the invented predicate encoding
with the main encoding by defining that
invented predicates can be used both as
head as well as body predicates in hypothesis rules.
Line~8 requires that arguments of invented predicates
are sorted lexically in the same way
as the IDs of their types are.
This reduces redundancy,
because it does not matter in practice
whether we use an invented predicate as
\verb+inv(Type1,Type2)+ or as \verb+inv(Type2,Type1)+,
as long as it is used in the same way in all rules.
Lines~9--11 perform further redundancy elimination
by defining which invented predicates are used,
and eliminating solutions where we guess the existence
of an invented predicate but do not use it.
Line~12 defines a cost for predicate invention in general,
line~13 defines a cost for each invented predicate,
lines~14--15 define extra cost if the invented predicate
is used both in the head and in the body of the rule in the answer set,
lines~16--18 define extra cost for multiple usages of the
same predicate in the rule body,
and lines~19--23 define a cost for pairs of \revA{distinct }invented predicates
where one is in the body and the other one in the head of a hypothesis rule:
cost is defined if the predicate in the head is lexicographically
greater or equal to the predicate in the body.
This prevents rules that can make cycles over
invented predicates early in the search process,
and intuitively \revA{prefers hypotheses
that are stratified \citep{Apt1988declknowledge}
with respect to invented predicates}.

\section{Best-Effort Learning and Prediction}
\label{secBestEffort}
{\revAc%
The \inspire\ system performs brave induction of explicitly given positive
and implicitly given negative examples,
which was sufficient for the competition.
The type of induction task that is solved
is similar to the task solved by the \xhail\ system.
Formally, \inspire\ searches for a hypothesis $H$
such that for each given example trace
$\langle \mi{trace}, \mi{valid\_moves} \rangle$
there is an $I \in \AS(bk \cup \mi{trace} \cup H)$
with $I$ containing the valid moves
specified in $\mi{valid\_moves}$
and no other valid moves.
}%

Algorithm~\ref{algo} shows the main algorithm
of the \inspire\ system
which is visualized from a conceptual point
of view in Figure~\ref{figFlowchart} (see page~\pageref{figFlowchart}).
The algorithm gets a competition instance (see Section~\ref{secILPCompetition})
as input:
background knowledge $bk$ is a set of ASP rules,
the set of examples $e$ is of form
$\mi{\langle trace, label \rangle}$
where \mi{trace} is a set of atoms for predicate \verb+agent_at+
and \mi{label} is a set of atoms for predicate \verb+valid_move+,
the mode bias $m$ is given in the form of
target predicate and relevant predicates,
and finally the set of test traces $\mi{tests}$
is of form $\mi{\langle trace, id \rangle}$
where \mi{trace} is a set of atoms for predicate \verb+agent_at+ and \mi{id}
is required for labeling prediction outputs
with the correct trace.

\figureAlgo
Initially, Algorithm~\ref{algo} sorts examples by length of their trace.
The variable $\mi{bestquality}$, initially zero,
stores the number of examples that we can predict
correctly with the best hypothesis found so far.
The loop in line~\ref{stepExLoop} iterates over the sorted
examples, starting with the smallest.
For each example,
the loop in line~\ref{stepCostLoop} iterates
over cost limit values from \climitmin\ to \climitmax.
\revA{For the competition, we set
$\climitmin \eqs 4$ and $\climitmax \eqs 15$,
in a general setting we use $\climitmin \eqs 1$ and $\climitmax \eqs \infty$.}
For each value of $\mi{climit}$,
in line~\ref{stepHypgen} we enumerate all answer sets
of $\Phypgen(m,\mi{climit})$ which denotes
the ASP encodings for hypothesis generation as described in
Figures~\ref{figCodeMain}, \ref{figCodeCost}, \ref{figCodeRed},
and~\ref{figCodeInv}.
\revA{%
The parameters $m$ and $\mi{climit}$ of encoding $\Phypgen$
are used as follows:
from the mode bias $m$,
facts are generated as described in Section~\ref{secHypgenInput},
and the value of $\mi{climit}$ is passed to ASP as a constant \texttt{climit}.}
In line~\ref{stepHspace}
each answer set is transformed into a (nonground) rule
which yields the hypothesis search space $\mi{hspace}$.

Given %
\mi{hspace},
we search for an optimal hypothesis
using the current example's
\mi{trace} and %
\mi{label}.
A hypothesis $h \subseteqs \mi{hspace}$
must predict the extension of %
\verb+valid_move+ in the label correctly
with respect to background knowledge $bk$ and \mi{trace}.
Note, that predicate \verb+valid_move+ is specific to the competition,
but our encodings are flexible with respect to
using different or even multiple predicates.
A correct hypothesis is optimal
if it has lower or equal cost
compared with all other correct hypotheses,
where cost is the sum of costs of rules used in the hypothesis
and cost of a single rule
is computed according to Section~\ref{secHypgenConfig}.

Hypothesis search is done using ASP optimization
on a program $P_\mi{hs}$
whose encoding \revA{is similar to}
the Inductive Phase encoding used in
\xhail~\cite[Section~3.3]{Ray2009}.
Briefly, $P_{hs}$ contains the background knowledge,
the current example's trace $\mi{trace}$ \revA{as facts},
a module $P_\mi{verify}(\mi{label})$
which eliminates solutions that do not satisfy the example,
{\revAc%
and a transformed rule $P_{\mi{rule}}(r)$
for each rule $r$ in the hypothesis space.
$P_\mi{verify}(label)$ contains
the set
$\{ \texttt{pos\_valid\_move($X,Y$).} \mid \texttt{valid\_move($X,Y$)} \in \mi{label} \}$
of facts which encode positive example traces;
a rule \quo{\texttt{covered :- $\mi{label}$.}}
which recognizes coverage of positive example parts;
and the following rules which recognize the entailment of the example
by checking the coverage of positive examples
and forbidding the violation of invalid moves:
\begin{equation*}
  \begin{array}{@{}l@{}}
  \texttt{violated :- valid\_move(X,Y), \naf\ pos\_valid\_move(X,Y).} \\
  \texttt{good\_example :- covered, \naf\ violated.} \\
  \texttt{:- \naf\ good\_example.}
  \end{array}
\end{equation*}
}%
$P_{\mi{rule}}(r)$ contains
\begin{inparaenum}[(i)]
\item the original rule $r$
  with an additional body condition \texttt{use($r$)};
\item
  \revA{a guess \texttt{\{ use($r$) \}.}
  which determines whether rule $r$ is part of the hypothesis};
  and
\item
  a weak constraint that incurs the cost of $r$
  if \texttt{use($r$)} is true.
\end{inparaenum}
\revA{In each answer set $I$ of $P_{hs}$,
the set of rules $\{ r \mid \texttt{use($r$)} \} \subseteqs \mi{hspace}$
is a hypothesis
that entails the given example $\langle\mi{trace}, \mi{label}\rangle$.
An optimal answer set of $P_{hs}$
is a hypothesis such that there is no cheaper hypothesis
$h' \subseteqs \mi{hspace}$
that entails the given example with smaller cost.}

If such a hypothesis $h$ exists,
in line~\ref{stepQuality} we measure the quality of $h$
by testing how many of the given examples $e$
are correctly predicted by $h$.
This test is performed by repeatedly evaluating
an ASP program on $bk$, $h$,
and the $\mi{trace}$ of the respective example.
If the obtained quality is higher than the best previously
obtained quality,
in lines~\ref{stepAttemptStart}--\ref{stepAttemptEnd}
we store the quality and the hypothesis
and make a prediction attempt for all test traces.
Prediction is performed by evaluating
an ASP program on $bk$, $h$, and on each trace.
If we have correctly predicted the labels
of all training examples $e$,
we immediately return the hypothesis
after the prediction attempt
(because we have no possibility to
measure hypothesis improvements beyond this point).
In case we do not find a hypothesis that entails all examples,
we return the hypothesis that entails most training examples.
If we do not find any hypothesis, we return \mi{null}.

For the competition,
it was only required that the system makes prediction attempts.
To provide a more general approach,
our algorithm also returns a hypothesis.

{\revAc
\subsection{Soundness and Completeness}
\label{secSoundComplete}

The \inspire\ system makes a best effort:
\begin{inparaenum}[(i)]
\item
  it learns hypotheses from a single example at a time,
\item
  it checks if the found hypothesis covers all examples,
\item
  it makes a prediction attempt if the found hypothesis
  covers more examples than the previously best hypothesis,
  and
\item
  if not all examples were covered,
  it continues searching for a hypothesis
  in those examples that were not used
  for hypothesis search so far.
\end{inparaenum}

The hypothesis search and optimization encoding of our system
is %
sound and complete for single examples,
that means every hypothesis that is returned will entail the respective example,
and the encoding will find all possible minimal hypotheses for a given example.
Algorithm~\ref{algo} ensures that the system
will only make a prediction based on a hypothesis
that can be verified on more examples than any previously found hypothesis.
This approach is a best effort, but it is not sound, as it returns solutions
that do not entail all examples.
It is also incomplete
because the system can find various cheap hypotheses for each example
while missing a more expensive hypothesis that entails all examples.
\emph{Soundness can be established} by returning only hypotheses
that cover all given examples
(which would reduce the score in the competition because
no partial credit would be gained).
Incompleteness is the main disadvantage of our system compared to \ilasp,
which becomes visible in our analysis of experiments on the instance level,
see Section~\ref{secInstanceByInstanceResults}.
Still, our system provides higher accuracy
in the competition settings by solving more instances
and by obtaining partial credit for hypotheses that entail only some examples.
In a setting where all examples can be represented in the same answer set
without interfering with one another,
for instance all ILP tasks that can be processed by \xhail,
\emph{completeness can be established}
by setting in line~\ref{stepPhs}
\begin{equation*}
P_{hs} := bk \cup
  \{ t \mid \langle t, l \rangle \in e \} \cup
  \{P_\mi{rule}(r) \mid r \in \mi{hspace}\} \cup
  \bigcup \{ P_\mi{verify}(l) \mid \langle t, l \rangle \in e \}
\end{equation*}
and by modifying Algorithm~\ref{algo} as follows:
remove the example loop (line~\ref{stepExLoop})
and the quality check (line~\ref{stepQuality}--\ref{stepRememberBestHyp} and line~\ref{stepCheckQuality}).
The first hypothesis that is found with this modified approach entails all examples.}

\section{Evaluation}
\label{secEvaluation}
The \inspire\ system
is \revA{implemented in }Python
and \revA{uses }the \revA{version~5.2.2 of the }ASP system \clingo\
which consists of the grounder
\gringo~\citep{Gebser2011gringo3}
and the solver \clasp~\citep{Gebser2012aij}.
\clingo\ is used for all ASP evaluations
shown in Figure~\ref{figFlowchart}.
\revA{%
The default configuration of this \clingo\ version provides
anytime answer sets during optimization
due to mixed usage of core-guided and model-guided optimization
\cite{Andres2012,Alviano2015aspopt}
in combination with stratification heuristics~\citep{Ansotegui2013maxsat}.
}%
\revA{%
Due to the short timeouts of the ILP competition,
we limited ASP computations to $T_\mi{lim} \eqs 5$ seconds each,
which was achieved by using the \clingo\ parameter
\texttt{-{}-time-limit=5}.
(See Table~\ref{tblResultsTimelimit} for experiments with $T_\mi{lim}$.)
}%

\subsection{Comparison System based on \ilasp}\label{secEvaluationILASP}
As there was no other participant in the competition,
and as the competition used a novel input format,
there was no state-of-the-art system we could use for comparison.
Therefore we adapted the state-of-the-art
\ilasp\ system~\citep{Law2017manual,Law2016ilasp2i} \revA{version 3.1.0},
using a wrapper script.
\revA{%
The wrapper converts target and relevant predicates
into mode bias commands.
Negative examples were specified implicitly in the competition:
if a move was not given as valid, it was invalid.
Therefore, the wrapper creates for each competition example one positive \ilasp\ example:
valid moves are converted to positive atoms,
trace atoms are converted into example context
(see \citep[Section~5]{Law2017manual}),
and all implicitly given invalid moves are converted into negative atoms.
}%
\begin{example}
  \revAc
  The trace and the valid moves from \texttt{\#Example(0)}
  in Figure~\ref{figSimpleTask}
  are converted into the following \ilasp\ example.
  \begin{LVerbatim}[gobble=2]
  #pos(ex0, { valid_move((0,1),0), valid_move((1,0),0), valid_move((1,1),1) },
            { valid_move((0,0),0), valid_move((1,1),0), valid_move((0,0),1),
              valid_move((0,1),1), valid_move((1,0),1) },
            { agent_at((0,0),0).  agent_at((0,1),1). }).
  \end{LVerbatim}
\end{example}
To increase \ilasp\ performance,
we generate \revA{negative example }atoms only for those time points
where an agent position exists in the trace
(most competition examples define 100 time points
but use less than 20).
We \revA{add the directives }%
\texttt{\#maxv(3)} and \texttt{\#max\_penalty(100)}
\revA{to configure the \ilasp\ bias;
these parameters were found in preliminary experiments:}
lower values did not yield any hypothesis
and higher values yielded much worse performance.
In summary, we did our best to make \ilasp\ perform well.

\revA{%
As \ilasp\ realizes two evaluation algorithms that are suitable for different kinds of instances
(Mark Law, personal communication),
we performed experiments with both algorithms:
by \ilaspm{2i} we refer to \ilasp\ with parameter \texttt{-{}-version=2i},
see \citep{Law2016ilasp2i},
and by \ilaspm{3} we refer to \ilasp\ with parameter \texttt{-{}-version=3},
see \citep{Law2017manual}.

We provide the \ilasp\ wrapper
at \texttt{https://bitbucket.org/knowlp/inspire-ilp-comp}
in directory \texttt{ilasp-wrapper}.
}%

\ResultsParameters
\subsection{Results}
\label{secResults}
We performed experiments on development set (D, 33 instances)
and on the test set (T, 45 instances)
of the ILP competition,
both available on the competition homepage \citep{ILPCompetition}.
Runs were performed for the resource limits of the competition (30~sec, 2~GB)
and for higher resource limits (600~sec, 5~GB).
\revA{Tables show averages over three independent runs
for each configuration. The tables contain}
the number of timeouts
(i.e., the system did not terminate within the time limit)
both absolute and in percent (lower is better);
the accuracy of the last attempt
for predicting test traces
(this value contains fractions if only some test traces
of an instance were predicted correctly);
the number of attempts performed,
and the average time $T$ and memory usage $M$.
\revA{%
Experimental results for the \inspire\ system
are shown for several cost settings
for negation and predicate invention,
see Table~\ref{tblParameters} for an overview.}

\Results
\revA{%
Table~\ref{tblResults} shows results comparing variations of \inspire\ and
\ilaspm{2i}.
With respect to \emph{timeouts},
we can see that a timeout of 600~sec yields
significantly fewer timeouts and higher accuracy
than the competition timeout of 30~sec
(both for \inspire\ and \ilasp).
A timeout occurs if no hypothesis is found
and no prediction attempt is made.
Therefore, it does not count as a timeout
when the \inspire\ system found a hypothesis
that predicted some given examples incorrectly and made a (best-effort)
prediction attempt on the test data.
While \ilasp\ can learn from multiple examples at once,
it does not support automatic predicate invention.%
\footnote{Invented predicates can manually be added
in an explicit specification of the search space,
however we provided \ilasp\ only with a mode bias.}
This explains the lower accuracy and the low number of prediction attempts
(\ilasp\ performs a prediction only if all examples are entailed by a hypothesis).
The \inspire\ system terminates without timeout for all instances
for configurations \inspireln\ and \inspireliln.
Due to learning from single examples,
and due to the upper limit \climitmax\ on hypothesis cost,
termination does not mean that all examples are covered,
therefore termination does not guarantee 100\% accuracy
(see also Section~\ref{secSoundComplete} about sound- and completeness).

With respect to \emph{accuracy},
the Test dataset appears to be more challenging than the Development
dataset.
\inspireln\ was used to participate in the competition,
and this configuration achieves the best accuracy for a time limit of 600~sec.
For the lower time limit, \inspireliln\
(reduced negation and reduced predicate invention)
yields the highest accuracy.
\emph{Attempts} are generally correlated with accuracy,
and it is visible that the limitation in this task is time, not memory.

To find the best configuration,
we compare accuracy using a one-tailed paired t-test
(paired because we perform experiments on the same instances).
\inspireliln\ and \inspireln\ are significantly better
than all other configurations with $p \lts 0.027$,
however comparing \inspireliln\ and \inspireln\ shows no significant difference
($p \eqs 0.1$).
Increasing predicate invention and negation
has the effect of more timeouts and lower accuracy.
We analyse results on an instance-by-instance basis in Section~\ref{secInstanceByInstanceResults}.
}%

\revA{%
Table~\ref{tblResultsIlasp} shows a comparison of \ilasp\ algorithms.
For the test set and high resources,
\ilaspm{3}\ yields slightly higher accuracy than \ilaspm{2i},
but overall, \ilaspm{2i}\ provides significantly better accuracy
than \ilaspm{3} with $p \lts 0.01$.
If we compare Development and Test instances
for a time budget of 600~sec,
we notice that both \ilasp\ algorithms
requires more time but less memory for the Test instances.
This shows that the Test set was structurally different
than the Development set,
see also Section~\ref{secInstanceByInstanceResults}.
}%

\ResultsIlasp
\revA{%
Table~\ref{tblResultsTimelimit}
shows experimental results for several $T_\mi{lim}$ values
and the default \inspire\ configuration.
The time limit $T_\mi{lim}$ has the primary purpose of providing some,
potentially suboptimal, solution within a limited time budget.
For participating in the competition, we used $T_\mi{lim}\eqs 5$.
The results in the table show that for the low-resource setting,
modifying this parameter has no effect on accuracy.
For the high-resource setting,
limiting the time yields significantly better accuracy
than omitting the timeout ($T_\mi{lim} \eqs \infty$)
with $p \lts 0.01$.
Note, that ASP Evaluation is performed multiple times in each run,
see Figure~\ref{figFlowchart} and Algorithm~\ref{algo},
therefore even a low time limit of 5~sec per ASP call
can yield an overall timeout of 600~sec.}

{\revAc
\subsubsection{Instance-by-instance analysis}
\label{secInstanceByInstanceResults}

The competition dataset comprises a development set of 11 instance types,
numbered 4 through~12, 14, and 17,
moreover there are 15 test instance types,
numbered 13, 15, 16, 18, 29, 30, 32 through 35, and 37 through 41.
Each instance type exists in three difficulty levels (easy, medium, and hard),
which yields a total of 33 Development and 45 Test instances.
In the following,
we refer to instance types unless otherwise noted.

In the experiments shown in Table~\ref{tblResults},
six instances (5, 14, 16, 29, 33, 40)
are solved neither by \inspire\ nor by \ilasp.
A close inspection shows that this is the case for various reasons.
One reason is inconsistency in the mode bias:
instance~5 contains atoms of form \texttt{link($\cdot,\cdot$)}
in background knowledge but not in the bias,
while the bias contains a predicate \texttt{full(cell)}
that does not exist in the background knowledge.
Another reason is,
that the time limit $T_\mi{lim}$ is too low:
instance~14 can be solved by \inspire\ with $T_\mi{lim} = 60$.

Ten instances (7, 9, 10, 11, 15, 16, 18, 32, 34, 38)
yield a (partially) correct prediction with \inspire\
and no prediction with \ilasp.
From these, four instances (7, 11, 32, 38)
require invented predicates in the hypothesis.
For two others,
\ilasp\ terminates without finding a hypothesis.
For the other four instances,
\ilasp\ exceeds the timeout.
Instance~38 can be solved with default parameters (\inspire)
but not with reduced predicate invention (\inspire$^{I\text{-}\text{-}}$),
while other instances yield results with both configurations.
This is an effect of the maximum cost limit setting \climitmax, which makes the approach incomplete if $\climitmax \lts \infty$, but yields better performance in the competition setting.
The successful hypothesis for Instance~38 comprises six rules
where three rules define an invented predicate
and one rule requires a negated invented predicate in the body.

\ResultsTimelimit
Three instances (17, 35, 39) are only solved by \ilasp.
Of these, instance~39 can be solved with \inspire\
with an increased timelimit $T_\mi{lim} = 60$.
The other instances require
learning from multiple examples at the same
to achieve a correct hypothesis.

For instances~15 and~17, only the hard version can be solved.
Hard versions of instances are extensions of easy version
with irrelevant mode bias instructions.
It seems that in these instances,
superfluous predicates made the induction problem easier to solve.

In summary,
several factors were important to solve competition instances
in a tight time budget,
most notably predicate invention, learning from multiple examples,
and choosing correct limits for the timeout $T_\mi{lim}$
of intermediate ASP evaluations.
}%

\section{Discussion}
\label{secDiscussion}
The \inspire\ system uses a hypothesis search space
of stepwise increasing complexity.
This is a commonly used approach in ILP
for investigating more likely hypotheses first.
Usually a very coarse-grained measure of rule complexity
is used such as the number of body atoms in a rule.
We extend this idea by using an ASP encoding
that provides a fine-grained, flexible, easily adaptable,
and highly configurable way of
describing hypothesis cost
for controlling the search space.
\revA{Our experiments show that the choice of cost parameters
has an influence on finding
hypotheses that generalize correctly within a given time budget.}

We support predicate invention.
\revA{%
Increasing the cost of predicate invention in the system (\inspirelili)
reduces the resource requirements of the system,
but unfortunately it also reduces the quality of hypotheses
and leads to lower accuracy on the Test dataset.
Likewise, increasing predicate invention (\inspiremi)
leads to lower accuracy because of timeouts that are caused
the search space growing too fast.
Hence, fine-grained configuration of predicate invention is important
for obtaining good results.
}

Invented predicates with arity one
are implicitly created by hypotheses containing
only reflexive usage of an invented predicate \texttt{inv},
i.e., hypotheses containing only \texttt{inv(V,V)}
for some variable \texttt{V}.
\revA{%
Having dedicated unary invented predicates would be a better solution,
however this would require to constrain binary invented predicates
such that they are used with two distinct arguments
in at least one rule head in the hypothesis
(otherwise they are effectively unary predicates).
Making a global constraint over the structure of the full hypothesis
is currently not possible in \inspire,
however an approach for such constraints has been described
by \citet{Athakravi2015ilpconstraintbias}.}

Enumerating answer sets of our hypothesis space generation
encodings in Algorithm~\ref{algo} line~\ref{stepHypgen}
is computationally cheap compared with
hypothesis search (line~\ref{stepHypSearch}).
\revA{%
For finding an optimal hypothesis,
the \inspire\ system uses an encoding which requires examples
to be represented in disjoint parts of the Herbrand Base,
see challenge (C1) in Section~\ref{secIntro}.
Therefore, we designed an algorithm that learns
from single examples at a time.
Our fine-grained hypothesis search space generation
(described in Section~\ref{secHypgen})
is independent from Algorithm~\ref{algo} and
it is compatible with other approaches for hypothesis search
that work with multiple examples and with noisy examples.
}%

The \inspire\ system learns from single examples
and sorts them by trace length,
which reduces resource consumption in the
hypothesis search step.
This is a strategy that is specific to the competition:
we observed that even short examples often provide
sufficient structure for learning the final hypothesis,
moreover competition examples were noiseless.
We think that future ILP competitions
should prevent success of such a strategy
by using instances that require learning from all examples
at once,
e.g., by providing smaller, partial, or noisy examples
(even in the non-probabilistic track).

Testing whether a hypothesis correctly predicts
an input example is computationally cheap.
Therefore, for each newly found hypothesis
we perform this check on all examples.
If this increases the amount of correctly predicted examples,
we make a prediction attempt on the test cases.
If all examples were correctly predicted,
we terminate the search, because
we have no metric for improving the hypothesis
after predicting all examples correctly
(competition examples are noiseless and our search
finds hypotheses with lower cost first).

A major trade-off in our approach is the blind search:
we avoid to extract hypotheses from examples
as done in the systems \xhail~\citep{Ray2009} and \iled~\citep{Katzouris2015}.
This means we rely on the mode bias and on our incrementally increasing
cost limit to obtain a reasonably sized search space for hypothesis search.

Note that we did not compare \inspire\ with
\xhail~\citep{Ray2009}, \mil~\citep{Muggleton2014metagol},
or \iled~\citep{Katzouris2015},
\revA{partially} because these approaches are not compatible with
challenge (C1)\revA{,
partially because these approaches are syntactically incompatible
with tuple terms (i.e., terms of form \texttt{cell((1,2))})
which are essential in the background knowledge of the competition.}

\section{Related Work}
\label{secRelated}

Inductive Logic Programming (ILP) is a multidisciplinary field and has been greatly impacted by Machine Learning (ML),
Artificial Intelligence (AI) and relational databases.
Several surveys such as those by \citet{Gulwani2015} and \citet{Muggleton2012ilp20} mention about ILP systems and applications of ILP in interdisciplinary areas.
Important theoretical foundations of ILP comprise
Inverse Resolution and Predicate Invention
\citep{muggleton1992machine,muggleton1995inverse}.

Most ILP research has been based on Prolog
and aimed at Horn programs that
exclude Negation as Failure which provides monotonic commonsense reasoning under incomplete information.
Recently, research on ASP-based
ILP \citep{Otero2001aspilp,Ray2009,Law2014ilasp}
has made ILP more declarative
(no necessity for cuts, unrestricted negation)
but also introduced new limitations
(scalability, predicate invention).
Predicate invention is indeed a distinguishing feature of ILP:
\cite{Dietterich2008} writes that
`without predicate invention, learning always will be shallow'.
Predicate invention enables learning an explicit representation
of a `latent' logical structure that is neither present in background knowledge nor in the input,
which is related to successful machine learning methods such as Latent Dirichlet Allocation \citep{Blei2003lda} and hidden layers in neural networks \citep{LeCun2015}.
\cite{Muggleton2015predinv} recently
introduced a novel predicate invention method
and, to the best of our knowledge for the first time,
compared \revA{implementations of Metagol in ASP and Prolog}.
Other ASP-based ILP solvers do not support
predicate invention,
or they support it only with an explicit specification
of rules involving invented predicates \citep{Law2014ilasp}.
In purely Prolog-based ILP,
several systems with predicate invention have been built
\citep{Craven2001,Muggleton1987duce,Flach1993},
however these systems also do not support the full power of
ASP-based ILP with examples in multiple answer sets \citep{Otero2001aspilp,Law2014ilasp}.
Note that predicate invention in general is still considered
an unsolved and very hard problem \citep{Muggleton2012ilp20}.

\revA{%
The TAL \citep{Corapi2010ilptal} ILP system
is based on translating an ILP task into an
Abductive Logic Programming instance \citep{Kakas1992},
where each rule in the search space
is represented by a ground atom that holds a list of body atoms and
a list of substitutions of mode bias placeholders with variables in the rule.
TAL is Prolog-based and does not instantiate all of these atoms at once.
The ASPAL \citep{Corapi2012ilpaspal} ILP system, similar to TAL,
represents rules as single atoms, however ASPAL replaces constants by unique variables
in order to instantiate the full search space,
followed by an ASP solver call that instantiates these special variables with constants
and finds the optimal hypothesis.
Our approach provides a declarative and configurable method
for enumerating they hypothesis space of an ILP problem.
We represent a single hypothesis rule not as a single atom
of form $\mi{rule}(\ldots)$ as done by TAL and ASPAL
but as a structured representation in a set of atoms in an answer set.
This has the advantage of reducing the size of the instantiation and enables
a fast enumeration of all hypothesis candidates.
The disadvantage of our approach is,
that after enumerating hypothesis candidates
we still need another approach for finding the best hypothesis,
whereas
TAL and ASPAL combine the search for hypothesis candidates
with the search for the optimal hypothesis in one representation.

The following extensions of ASPAL,
which impose restrictions on hypotheses,
have important similarities with our approach.
ASPAL was extended in order to perform
\emph{minimal revision} of a logic program
using ILP \citep{Corapi2011normativedesignilp}
by giving hypotheses in the bias a cost according to
their edit distance to a given logic program.
Revising a logic program was applied to solving ILP tasks
in another extension \citep{Athakravi2013ilphyprefasp} of ASPAL:
initially, a hypothesis that partially covers the examples and has limited complexity
is learned,
followed by revisions, called \emph{change transactions},
of limited complexity which aim to entail more examples by modifying the hypothesis.
This extension of ASPAL limits the search space relative to an intermediate hypothesis,
which can rule out the empty hypothesis as solution of certain intermediate steps.
Opposed to that, our approach never puts a lower limit on the hypothesis complexity:
in \inspire, the search space always increases with the number of iterations,
while change transactions have the potential to generate big hypotheses
based on a sequence of comparatively small searches for locally optimal ILP solutions.
ASPAL was extended with a \emph{constraint-driven} bias
\citep{Athakravi2015ilpconstraintbias}
where the hypothesis search space,
usually given only by a set of head and body mode declarations,
can be constrained in a more fine-grained manner.
A set of domain-specific rules and constraints
(e.g., to require that hypothesis rules with a specific predicate in the head
must contain another specific predicate in the body)
or generic constraints (e.g., to require that the hypothesis is a stratified program)
imposes hard constraints on hypotheses
and these constraints can be formulated differently for each predicate in the mode bias.
Different from the constraint-driven bias,
we define a preference function and merely delay,
but do not completely exclude,
hypothesis candidates from the search.
Our approach is controlled by cost coefficients
and does not permit domain-specific formulation of preferences.
Both approaches could be combined
to obtain (i) domain-specific control over the overall search space
as well as (ii) domain-specific control over search space extension
(in case no solution is found in the initial search space).
This could be achieved by extending the ASPAL constraint-driven bias language
with weak constraints.}

A recent application of ASP-based ILP was done by
\cite{mitra2016addressing},
who perform Question Answering on natural language texts.
Based on statistical NLP methods,
they gather knowledge and learn learning with ILP
how to answer questions
similar to a given training set.
They used \xhail~\citep{Ray2009} to learn
non-monotonic hypotheses in a
formalization of an agent theory with events.

\section{Conclusion}
\label{secConclusion}
We created the \inspire\ Inductive Logic Programming system
which supports predicate invention
and generates the hypothesis search space
from the given mode bias using an ASP encoding.
This encoding provides many parameters for a
fine-grained control over the cost of rules
in the search space.

\revA{%
It is useful to have a fine-grained control over the order
in which the search space is explored,
in particular for controlling negation and predicate invention.
Invented predicates are more often useful for abstracting from other concepts,
and less often useful if they generate answer sets by introducing additional non-determinism.
Similarly, abundance of negation will easily introduce
(potentially) undesired non-determinism within the hypothesis.
For the ILP competition,
the \inspire\ system was configured to first explore hypotheses
with mainly positive body literals
(see parameter \mb{cost\_negbodyliteral})
and hypotheses where invented predicates
are used in a stratified way
(see parameter \mb{cost\_inv\_headbodyorder}).}

The performance of our system is provided by
\begin{inparaenum}[(i)]
\item
  appropriately chosen cost parameters
  for the shape of rules in the hypothesis search space,
\item
  incremental exploration of the search space,
\item
  an algorithm that learns from a single example at a time,
  and
\item
  the usage of modern ASP optimization techniques
  for hypothesis search.
\end{inparaenum}
On ILP competition instances, \inspire\ clearly outperforms
\ilasp, which we adapted to the competition instance format.

The \inspire\ system was created specifically for the first ILP competition,
but the fine-grained control over the hypothesis search space generation
is \revA{a generic method that is }%
independent from the learning algorithm
and could be integrated into other systems,
for example into \revA{\xhail\ or }\ilasp.

Hypothesis candidates are generated in a blind search,
i.e., independent from examples.
This might seem like a bad choice,
however it is a viable option in ASP-based ILP
because we found in recent research \citep{Kazmi2017xhailchunking}
that existing methods for non-blind search
have major issues which make their usage problematic:
the \xhail~algorithm \citep{Ray2009}
produces many redundancies in hypothesis generation,
leading to a very expensive search (Induction),
while the \iled~algorithm \citep{Katzouris2015} is unable
to handle even small inconsistencies in input data,
leading to mostly empty hypotheses or program aborts.

We conclude that the good performance of the \inspire\ system
is based partially on our novel fine-grained hypothesis
search space generation,
and partially on the usage of peculiarities of competition instances.
To advance the field of Inductive Logic Programming
for Answer Set Programming,
future competitions should use more diverse instances
that disallow finding solutions from single examples,
while at the same time requiring the hypothesis
for separate examples to be entailed in separate answer sets.
To avoid blind search,
it will be necessary to improve algorithms and systems
to make them resistant to noise
and scalable in the presence of
large amounts of training examples.
We believe that theoretical methods developed
in Prolog-based ILP,
for example \citep{Muggleton2015predinv},
will be important and useful
for advancing ASP-based ILP.

\revA{%
As future work,
our hypothesis space generation approach could be integrated into
an existing ILP system, for example into the open source \xhail\ system
by modifying the method \texttt{getKernel()}
in class \texttt{xhail.core.entities.Grounding}.
Note, that this would not make \xhail\ compatible with examples
of the ILP competition due to challenge (C1).
Another future work would be to make the \inspire\ input format more generic
and to replace the hypothesis optimization encoding with an approach
that can process multiple examples at once and noisy examples,
e.g., with the full \xhail\ encoding or with the encoding
from \ilasp\ version 1 \citep{Law2014ilasp}.}

The \inspire\ system and the \ilasp\ wrapper
are open source software and publicly available at
{\tt https://bitbucket.org/knowlp/inspire-ilp-comp}%
\enspace{}.

\section*{Acknowledgements}

This work has been supported by The Scientific and Technological Research Council of Turkey (TUBITAK) under grant agreement 114E777,
by the Austrian Science Fund (FWF) under grant agreement P27730,
and by the Austrian Research Promotion Agency (FFG) under grant agreement 861263.

\ifinlineref
\input{references.sty}
\else
\bibliography{library-backup}
\fi

\end{document}